\documentclass[letterpaper, 10 pt, conference]{ieeeconf}  

\IEEEoverridecommandlockouts                              

\usepackage[table,xcdraw]{xcolor}
\usepackage{etoolbox} 
\usepackage[utf8]{inputenc} 
\usepackage[T1]{fontenc}    
\usepackage[english]{babel} 
\usepackage[protrusion=true,expansion=true]{microtype}
\usepackage{comment}
\usepackage{cancel}
\usepackage{csquotes}
\usepackage{listings}
\usepackage{nameref}
\usepackage{paralist}
\usepackage{xspace}
\usepackage{setspace}
\usepackage{makeidx}
\usepackage{pdfpages}
\usepackage{footnote}

\usepackage{hyperref}
\hypersetup{
    colorlinks=true,
    linkcolor=black,
    citecolor=black,
    filecolor=cyan,
    urlcolor=blue
}

\usepackage{flushend}
\usepackage{balance}
\usepackage{anyfontsize}

\usepackage{paralist}
\let\labelindent\relax 
\usepackage{enumitem}
\setlist[itemize]{noitemsep,leftmargin=*,topsep=0in}
\setlist[enumerate]{noitemsep,leftmargin=*,topsep=0in}

\usepackage{amsmath,amssymb} 
\usepackage{amsfonts}       
\usepackage{mathtools}
\usepackage{array} 
\usepackage{nicefrac}       
\usepackage{breqn}          
\usepackage{textcomp}
\usepackage{bm} 
\usepackage{pifont}
\usepackage[
  separate-uncertainty = true,
  multi-part-units = repeat
]{siunitx}

\usepackage{graphicx}
\usepackage{adjustbox} 
\usepackage{epsfig}

\usepackage{float}
\usepackage{subcaption}
\usepackage[font=small,labelfont=bf]{caption}
\usepackage{lscape}      
\usepackage{wrapfig}

\usepackage{tikz}  
\usepackage{makecell}
\usepackage{overpic}
\usepackage{algorithm}
\usepackage[noend]{algpseudocode}

\usepackage{tabularx}
\usepackage{multirow}
\usepackage{multicol}
\usepackage{booktabs}   
\usepackage{array} 

\usepackage{longtable}
\usepackage{threeparttable}

\makesavenoteenv{tabular}
\makesavenoteenv{table}

\makeatletter
\let\NAT@parse\undefined
\makeatother
\usepackage[numbers,sort&compress]{natbib}

\usepackage{titlesec}
\titlespacing{\section}{0pt}{0.3\baselineskip}{0.25\baselineskip}
\titlespacing{\subsection}{0pt}{0.2\baselineskip}{0.15\baselineskip}
\titlespacing{\subsubsection}{0pt}{0.05\baselineskip}{0.03\baselineskip}

\renewcommand{\paragraph}[1]{\vspace{0.2em}\noindent\textit{#1} --}

\newcommand\mybar{\kern1pt\rule[-\dp\strutbox]{.8pt}{\baselineskip}\kern1pt}

\newcommand*\colourcheck[1]{%
  \expandafter\newcommand\csname #1check\endcsname{\textcolor{#1}{\ding{52}}}%
}
\colourcheck{green}

\newcommand*\colourx[1]{%
  \expandafter\newcommand\csname #1x\endcsname{\textcolor{#1}{\ding{55}}}%
}
\colourx{red}

\newcommand{\Orbit}{\textsc{Orbit}\xspace}
\newcommand{\simName}{\textsc{Orbit}-Surgical\xspace}

\usepackage{titlesec}

\title{\LARGE \bf
 \simName: An Open-Simulation Framework for \\ Learning Surgical Augmented Dexterity
}

\author{
Qinxi Yu$^{*,1}$,
Masoud Moghani$^{*,1}$,
Karthik Dharmarajan$^{2}$,
Vincent Schorp$^{2,3}$,
William Chung-Ho Panitch$^{2}$, \\
Jingzhou Liu$^{1,5}$,
Kush Hari$^{2}$,
Huang Huang$^{2}$,
Mayank Mittal$^{3,5}$,
Ken Goldberg$^{2}$,
Animesh Garg$^{1,4,5}$
\thanks{*These authors contributed equally to this work}
\thanks{$^{1}$University of Toronto, $^{2}$University of California, Berkeley, $^{3}$ETH Zurich, $^{4}$Georgia Institute of Technology, $^{5}$NVIDIA}%
\thanks{Correspondence to: \href{mailto:moghani@cs.toronto.edu}{moghani@cs.toronto.edu}, \href{mailto:garg@cs.toronto.edu}{garg@cs.toronto.edu}}%
}

\let\oldtwocolumn\twocolumn
\renewcommand\twocolumn[1][]{%
    \oldtwocolumn[{#1}{
    \begin{center}
    \vspace{-.2in}
    \includegraphics{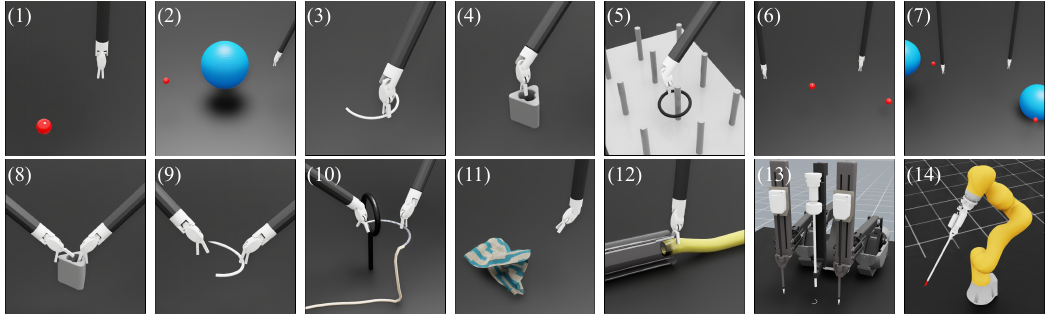}
    \captionof{figure}{\textbf{\simName Simulation Benchmark Tasks.} (1) \texttt{Reach}: dVRK Patient Side Manipulator (PSM) to reach a desired position (red sphere), (2) \texttt{Reach with Obstacles}: reach a desired position (red sphere) with randomly placed obstacles in the scene (blue sphere objects; object shape and size are customizable), (3) \texttt{Suture Needle Lift}: lift a suture needle to a desired position, (4) \texttt{Peg Block Lift}: lift a peg block to a desired position, (5) \texttt{Pick and Place}: pick and place a ring on a peg tower, (6) \texttt{Dual-arm Reach}: dual-arm reach to specific desired positions shown with red spheres, (7) \texttt{Dual-arm Reach with Obstacles}: dual-arm reach to specific desired positions (red spheres) with randomly placed obstacles in the scene, (8) \texttt{Pick and Transfer}: pick and transfer a peg block, (9) \texttt{Needle Handover}: handover and regrasp a suture needle, (10) \texttt{Threaded Needle Pass Ring}: handover a threaded suture needle through a ring pole, (11) \texttt{Gauze Cloth Pick}: retrieve a gauze and lift it to a desired location, (12) \texttt{Shunt Insertion}: insert a shunt (yellow tube) into a blood vessel phantom (clear tube), (13) \texttt{Multi-arm dVRK}: needle handover task with camera input from the dVRK Endoscopic Camera Manipulator (ECM), (14) \texttt{STAR Reach}: STAR arm to reach a desired position.
    \\
    Best viewed in color. Please see task videos at \href{https://orbit-surgical.github.io/}{orbit-surgical.github.io}}
    \label{fig:fig1}
    \end{center}
    }]
}

\begin{document}
\maketitle
\thispagestyle{empty}
\pagestyle{empty}

\begin{abstract}
Physics-based simulations have accelerated progress in robot learning for driving, manipulation, and locomotion. Yet, a fast, accurate, and robust surgical simulation environment remains a challenge. In this paper, we present \textbf{\simName}, a physics-based surgical robot simulation framework with photorealistic rendering in NVIDIA Omniverse. We provide 14 benchmark surgical tasks for the da Vinci Research Kit (dVRK) and Smart Tissue Autonomous Robot (STAR) which represent common subtasks in surgical training. \simName leverages GPU parallelization to train reinforcement learning and imitation learning algorithms to facilitate study of robot learning to augment human surgical skills. \simName also facilitates realistic synthetic data generation for active perception tasks. We demonstrate \simName sim-to-real transfer of learned policies onto a physical dVRK robot.

\noindent Project website: \href{https://orbit-surgical.github.io/}{orbit-surgical.github.io} 
\end{abstract}

\section{Introduction}

Robotic surgical assistants (RSA)  have been widely adopted since the early 2000s \cite{onnasch2002five, tewari2002technique, bodner2004first}. RSAs such as Intuitive Surgical’s da Vinci surgical systems enable minimally invasive robotic-assisted surgery, allowing surgeons to perform procedures using fewer, smaller incisions, reducing the risk of complications and recovery time. With enhanced capabilities and prevalence of RSAs, there is an increasing interest in augmented surgical dexterity~\cite{kyg} for partial automation of specific, monotonous, and error-prone subtasks in surgical procedures. Augmented surgical dexterity has potential to decrease cognitive load, reduce the learning curve, and improve patient outcomes.

Progress in augmented surgical dexterity has evolved slower than applications for robotic manipulation~\cite{allshire2022transferring, handa2022dextreme, mahler2019learning}, locomotion~\cite{rudin2022learning}, and autonomous driving~\cite{kaur2021survey}. These advancements in robot learning often depend on the utilization of simulation environments for either synthetic data generation or interaction based learning~\cite{todorov2012mujoco,coumans2016pybullet,makoviychuk2021isaac,mittal2023orbit}. 
While autonomous driving has progressed to Level 3 autonomy and is approaching Level 4, the majority of robotic surgical platforms still lack any form of autonomy~\cite{attanasio2021autonomy}.

A number of research efforts towards learning-based RSAs have led to simulating surgical subtasks~\cite{bourdillon2023integration,scheikl2023lapgym,xu2021surrol,varier2022ambf, tagliabue2020soft,richter2019open}. Yet, they lack accelerated physics to enable learning long-horizon manipulations. This results in a high barrier for researchers to contribute to the field, and impedes research for algorithmic development. Surgical environments are difficult to simulate due to the complicated interactions of rigid \& deformable components and lack of physics simulation models. Additional challenges include motion restriction by the remote center of motion (RCM) in trocars for laparoscopic surgery, and simultaneous control of multiple surgical instruments~\cite{scheikl2023lapgym}. Moreover, sensor simulation to achieve realistic endoscopic images also remains elusive.

\begin{table*}
\centering
\footnotesize
\caption{\textbf{Comparison Between Free Open-source Surgical Robotic Simulation Frameworks Available for Robot Learning.}}
\label{table:table1}
\begin{threeparttable}
\begin{tabular}{llllcccccc}
\toprule
\rowcolor[HTML]{D7DAFF} 
 &  &  & Collider & GPU & \multicolumn{4}{c}{Supported Dynamics} & Active \\ \cline{6-9} 
\rowcolor[HTML]{D7DAFF} 
\multirow{-2}{*}{\cellcolor[HTML]{D7DAFF}Name} & \multirow{-2}{*}{\cellcolor[HTML]{D7DAFF}Physics Engine}  & \multirow{-2}{*}{\cellcolor[HTML]{D7DAFF}Renderer}  &  Approximation & Acceleration & Rigid & Soft & Cloth & Fluid & Development \\

\midrule \midrule

LapGym \cite{scheikl2023lapgym} & SOFA & OpenGL & Surface meshes & \redx\tnote{*}& \greencheck\tnote{*} & \greencheck & \greencheck & \redx & \greencheck \\
\rowcolor[HTML]{EFEFEF} 
SurRol \cite{xu2021surrol} & Bullet~\cite{coumans2016pybullet} & OpenGL & Convex decomposition & \redx & \greencheck & \redx & \redx & \redx & \greencheck \\
AMBF-RL \cite{varier2022ambf} & Bullet~\cite{coumans2016pybullet} & OpenGL & Surface meshes & \redx&  \greencheck & \greencheck & \redx & \redx & \redx \\
\rowcolor[HTML]{EFEFEF} 
UnityFlexML \cite{tagliabue2020soft} & Flex~\cite{flex} & Unity3D & Not specified & \redx & \redx & \greencheck & \redx & \redx & \redx \\
dVRL \cite{richter2019open} & V-REP \cite{fontanelli2018v} & OpenGL & \redx & \redx & \greencheck & \redx & \redx & \redx & \redx \\
\rowcolor[HTML]{EFEFEF} 
\cellcolor[HTML]{EFEFEF} & \cellcolor[HTML]{EFEFEF} & Photorealistic & SDF & \cellcolor[HTML]{EFEFEF} & \cellcolor[HTML]{EFEFEF} & \cellcolor[HTML]{EFEFEF} & \cellcolor[HTML]{EFEFEF} & \cellcolor[HTML]{EFEFEF} & \cellcolor[HTML]{EFEFEF} \\
\rowcolor[HTML]{EFEFEF} 
\cellcolor[HTML]{EFEFEF} & \cellcolor[HTML]{EFEFEF} & Ray-tracing & Convex decomposition & \cellcolor[HTML]{EFEFEF} & \cellcolor[HTML]{EFEFEF} & \cellcolor[HTML]{EFEFEF} & \cellcolor[HTML]{EFEFEF} & \cellcolor[HTML]{EFEFEF} & \cellcolor[HTML]{EFEFEF} \\
\rowcolor[HTML]{EFEFEF} 
\multirow{-3}{*}{\cellcolor[HTML]{EFEFEF}Surgical Gym \cite{schmidgall2023surgical}} & \multirow{-3}{*}{\cellcolor[HTML]{EFEFEF}PhysX 5.1} & (OV-RTX) & Convex hull & \multirow{-3}{*}{\cellcolor[HTML]{EFEFEF}\greencheck} & \multirow{-3}{*}{\cellcolor[HTML]{EFEFEF}\greencheck} & \multirow{-3}{*}{\cellcolor[HTML]{EFEFEF}\redx} & \multirow{-3}{*}{\cellcolor[HTML]{EFEFEF}\redx} & \multirow{-3}{*}{\cellcolor[HTML]{EFEFEF}\redx} & \multirow{-3}{*}{\cellcolor[HTML]{EFEFEF}\greencheck} \\

\midrule

\multirow{3}{*}{\simName} & \multirow{3}{*}{PhysX 5.3} & \multirow{3}{*}{\begin{tabular}[l]{@{}l@{}}Photorealistic\\Ray-tracing\\ (OV-RTX) \end{tabular}} & \multirow{3}{*}{\begin{tabular}[l]{@{}l@{}}SDF\\ Convex decomposition\\ Convex hull\end{tabular}} & \multirow{3}{*}{\greencheck} &  \multirow{3}{*}{\greencheck} & \multirow{3}{*}{\greencheck} & \multirow{3}{*}{\greencheck} & \multirow{3}{*}{\greencheck} & \multirow{3}{*}{\greencheck} \\
 &  &  &  &  &  &  &  \\
 &  &  &  &  &  &  &  \\
\bottomrule
\end{tabular}
\begin{tablenotes}
\item[*] The check (\greencheck) and cross (\redx) denote presence or absence of the feature.
\end{tablenotes}
\end{threeparttable}
\end{table*}

We present \textbf{\simName}, an open-source surgical robot simulator, built on top of NVIDIA Isaac Sim~\cite{mittal2023orbit}. It provides stable GPU-accelerated physics with ray-traced rendering, rich suite of rigid and deformable object models, as well as a standardized, intuitive, and modular programming interface. \simName supports a variety of robot learning workflows: reinforcement learning (RL) and imitation learning (IL), teleoperation, and synthetic data generation. \simName currently supports two surgical robotic platforms; da Vinci Research Kit (dVRK)~\cite{kazanzides2014open} and Smart Tissue Autonomous Robot (STAR)~\cite{shademan2016supervised}. \simName includes simulated surgical subtasks intended for basic surgical maneuver training, along with examples of rigid and deformable surgical materials. The environments are modular to enable the robot learning community to quickly evaluate new algorithms on detailed surgical scenes. Summary of contributions:

\begin{enumerate}[
    topsep=0pt,
    leftmargin=*
    ]
    \item \textbf{GPU-based surgical simulation:} An open-source surgical robotics simulator with GPU-accelerated physics, contact-rich physical interaction, and high-fidelity rendering, extending  \Orbit~\cite{mittal2023orbit} framework for robotic surgery.
    
    \item \textbf{Surgical task benchmarks}: The simulation supports realistic interactions with both rigid and deformable materials, a variety of standard implementations of RL and IL algorithms, and various input peripherals including keyboard, spacemouse, gamepad, VR controller, and dVRK Master Tool Manipulator (MTM) for human-expert demonstration collection.

    \item \textbf{Simulation experiments:} Experiments using RL and IL methods demonstrating proof-of-concept policies for the provided surgical tasks. We also evaluate sub-policy learning for long-horizon surgical robotic tasks.

    \item \textbf{Synthetic data generation:} We evaluate \simName for generative synthetic images to train a needle segmentation framework with a combination of real and synthetic data to improve perception performance by over 2$\times$.
     
    \item \textbf{Real robot experiments:} Experiments demonstrating sim-to-real transfer of state-based and RL policies from \simName to a physical dVRK platform, showing transferability for real-world policy deployment.
\end{enumerate}

\begin{figure}
    \centering
    \includegraphics[width=\linewidth]{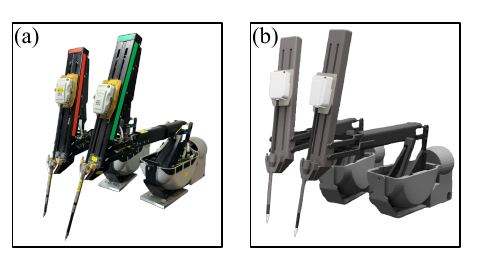}
    \caption{(a) Physical da Vinci Research Kit (dVRK) platform, (b) \simName simulation.} 
    \label{fig:fig2}
\end{figure}

\begin{table}
\centering
\footnotesize
\caption{\textbf{Frame Rate Comparison Between Surgical Robotic Simulation Frameworks.} The frame rate in frames per second (FPS) for different simulators are evaluated on three environments: robot reach task (Robot), rigid object manipulation (Rigid), and soft object manipulation (Soft). We evaluate all the environments based on the default environment configurations, and the throughput is calculated based on 100 simulation steps. For consistency in hardware comparison, only simulators compatible with the Ubuntu 20.04 LTS are assessed.}
\begin{tabular}{lccc}
\toprule
\rowcolor[HTML]{D7DAFF}
Framework & Robot & Rigid & Soft \\
\midrule
\midrule
SurRol \cite{xu2021surrol} & 129 & 92 & N.A. \\ 
\rowcolor[HTML]{EFEFEF} 
LapGym \cite{scheikl2023lapgym} & 621 & 617 & 77 \\ 
\midrule
\simName & 189732 & 87805 & 2048 \\ 
\rowcolor[HTML]{EFEFEF} 
Speedup & \textbf{305$\times$} & \textbf{142$\times$} & \textbf{27$\times$} \\
\bottomrule
\end{tabular}
\label{table:table2}
\end{table}

\section{Related Work}

\noindent \textbf{Simulators in Robot Learning}
Several robot learning efforts have introduced domain specific frameworks, often catering to specialized needs. Many prior frameworks~\cite{zhu2020robosuite,james2019rlbench} using MuJoCo~\cite{todorov2012mujoco} or Bullet~\cite{coumans2016pybullet} focus mainly on rigid object manipulation tasks. On the other hand, frameworks for deformable bodies~\cite{antonova2021dynamic, Lin2020softgym} mainly employ Bullet~\cite{coumans2016pybullet} or FleX~\cite{flex}, which use particle-based dynamics for soft bodies and cloth simulation. Most of these physics engines are CPU-based, relying on CPU clusters for parallelization. However, limited tooling exists for unified frameworks to enable content development specific to domains in surgery. \simName builds on recently released framework \Orbit, which provides a modular and unified simulation interface for robot learning. \simName relies on GPU-accelerated physics engines, signed-distance field (SDF) collision checking, and stable solvers based on FEM for deformable body simulation~\cite{neohooken2021miles, xpbd2019miles}.

\noindent \textbf{Learning in Surgical Robotics}
Robot learning for surgical tasks has been attempted for a number of subtasks with varying levels of autonomy~\cite{nagy2020autonomous,ficuciello2019autonomy,attanasio2021autonomy} such as shape cutting~\cite{murali2015learning,thananjeyan2017multilateral}, suturing~\cite{sen2016automating,krishnan2017transition}, debridement~\cite{murali2015learning}, as well as dissection and tissue retraction~\cite{nagy2020autonomous}. However, learning at scale has been limited due to lack of easy to use, high quality simulation frameworks for surgical robotics. Prior state-of-the-art surgical robotics simulators have suffered from issues such as limited rendering capabilities \cite{scheikl2023lapgym}, few available surgical environments \cite{varier2022ambf, tagliabue2020soft}, and lack of native support for deformable objects \cite{xu2021surrol}. But most of all, the lack of GPU-accelerated physics prevents learning~\cite{munawar2019real}. Further, there are numerous proprietary systems such as SimNow~\cite{simnow} and Mimic \& Simbionix~\cite{ss-mimic}. These offer virtual environments primarily for teleoperated skill and procedural practice, while lacking programmatic access for learning based methods, and high-speed throughput. As a result, a surgical robotic framework that integrates precise physics, delivers high speed realistic rendering, and accommodates various robot learning tools continues to be an ongoing challenge. TABLE~\ref{table:table1} provides a comparative overview of open simulation frameworks for surgical robotics. In addition, we compare the throughputs of different simulators on environments with similar complexity; namely, robot reach task, rigid object manipulation, and soft object manipulation. We provide the results in TABLE~\ref{table:table2}.

\section{\simName}

\subsection{Framework Overview}

\simName uses NVIDIA Isaac Sim and builds on the \Orbit framework enabling fast and accurate physics interactions, scene rendering, as well as access to robot learning libraries geared toward surgical subtask learning. \simName supports parallelized GPU simulation \cite{makoviychuk2021isaac}, contact-rich interactions (i.e. SDF-based collision meshes \cite{narang2022factory} and convex decomposition \cite{mamou2016volumetric}), and APIs for object definition (i.e. rigid body, FEM-based deformable body, and well as particle based simulation).

\noindent \textbf{Simulation Environment Design:} The framework architecture consists of a world and an agent, similar to how an operating system runs on a real-world robot to allow interaction with its environment. The agent receives observations from the world (i.e. environment), and computes actions to be applied on the surgical robot to interact with the world accordingly. Here, the world is defined as an entity consisting of the robot, sensory elements, and surrounding objects. The world description is modular in its core and can be designed in standardized scripts or through the graphical interface of NVIDIA Isaac Sim. This modularity and flexibility will allow the community to easily alter and introduce new elements into the existing world environments to meet the needs and features of their surgical applications. The surgical robot is the main component of the simulated surgical scenes. It embodies realistic joint articulations, actuator models, controllers, and collision properties to enable rich interactions with the environment and facilitate the sim-to-real transfer of control policies onto real-world robots.

\noindent  \textbf{Controllers:} We inherit GPU-based implementations of differential inverse kinematic and joint-level controllers from the \Orbit framework. The inverse kinematic models determine the joint positions needed to achieve the desired gripper pose in the Cartesian space. The inverse kinematic methods used to compute the inverse of the Jacobian include Moore-Penrose pseudo-inverse, adaptive SVD, Jacobian transpose, as well as damped least square (Levenberg-Marquardt).

\noindent  \textbf{Action Space:} We support both the Cartesian-space control and the joint-space control as our action space, enhancing the adaptability and transferability of the trained policies to a real-world dVRK platform. In Cartesian-space, the orientation of the gripper can be represented either as Euler angles or quaternions. Therefore, the quaternion-based action space involves ($d_x$, $d_y$, $d_z$, $q_{w}$, $q_{x}$, $q_{y}$, $q_{z}$, $d_g$), where $d_x$, $d_y$, $d_z$ determine the position, $q_{w}$, $q_{x}$, $q_{y}$, $q_{z}$ determine the orientation of the gripper in the Cartesian space, and $d_g$ determines if the gripper jaws are open/closed, respectively. The dVRK Patient Side Manipulator (PSM) has seven degrees-of-freedom (DOF). The first six DOFs correspond to the arm movement and the last DOF determines the gripper state. The arm DOFs consist of revolute (R) and prismatic (P) joints articulated in RRPRRR sequence \cite{xu2021surrol}. In joint-space mode, the action space is defined as ($q_{1}$, $q_{2}$, $q_{3}$, $q_{4}$, $q_{5}$, $q_{6}$, $d_g$) where $q_{1}$, $q_{2}$, $q_{3}$, $q_{4}$, $q_{5}$, $q_{6}$ determine the joint positions of the  arm, and $d_g$ determines the state of the gripper jaws, respectively.

\begin{figure*}
    \centering
    \begin{minipage}[c]{0.57\textwidth}
        \includegraphics[width=\textwidth]{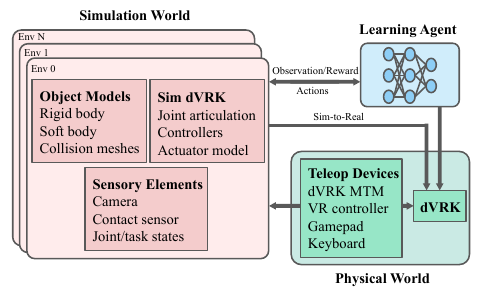}    
    \end{minipage}
    \,
    \begin{minipage}[c]{0.41\linewidth}
    \caption{\textbf{\simName Framework Architecture.} The \Orbit framework comprises of a simulation world, intelligent agents, and transfer of intelligence trained in simulation to the real world. The \simName simulation world includes realistic clones of surgical robots with joint articulation and low-level controllers. The object models include a rich suite of rigid and soft objects with precise collision properties. Sensory elements in the simulation imitate the real world sensors to provide training data. Peripheral I/O devices including the dVRK MTM are connected to the simulation environment to enable teleoperation of the simulated surgical robots. This setup facilitates the real-time capture of inputs from human experts, leveraging real-world demonstrations for policy learning in simulation.} 
    \end{minipage}
    \label{fig:fig3}
\end{figure*}

\noindent  \textbf{Observation Space:} We support working with diverse observations methods including the ground-truth states from the simulation environment (e.g. 6D pose of the gripper, object position in the Cartesian space) as well as various sensor simulations (e.g. cameras, height scanners, range sensors, contact sensors). The observation modalities are defined according to the specific needs of each task environment. In this work, we mainly focus on state-based observations for surgical skill learning to provide benchmark results.

\noindent  \textbf{Object Assets:} The dVRK robot is based on meshes from AMBF \cite{munawar2019real} and SurRol \cite{xu2021surrol} projects. The STAR arm is from Surgical Gym \cite{schmidgall2023surgical}. Surgical assets are partially created in Blender and partially from SurRol \cite{xu2021surrol}. The meshes are converted to USD file format in NVIDIA Isaac Sim. USD files include material definitions, geometric representations, and collision meshes among other asset attributes.

\subsection{\simName: Task Benchmark}

\simName provides surgical training task benchmarks (see Fig. \ref{fig:fig1}) inspired by Fundamentals of Laparoscopic Surgery (FLS) training curriculum \cite{fls} to reflect the technical skills for basic surgical maneuvers. Each task is provided in a standalone environment for benchmarking. These environments can be used to generate hand-scripted trajectories, collect human-expert data using multiple input peripherals, as well as training RL and IL policies to facilitate advancements in surgical autonomy. Up to 8000 environments can be simulated in parallel on a single NVIDIA RTX 3090 GPU for efficient data collection and fast surgical robot learning. Environments are modular and can be modified or integrated with other robots or tasks. An illustration of each task along with a short description is provided in Fig.~\ref{fig:fig1}.

\subsection{Robot Learning Workflows}

\textbf{Reinforcement Learning:} We support various reinforcement learning (RL) frameworks; rl-games \cite{rl-games2021}, RSL-rl \cite{rudin2022learning}, stable-baselines-3 (SB3) \cite{raffin2021stable}. In this work, we mainly utilise the RSL-rl library, as it is optimized for vectorized simulation and GPU training. We use the Proximal Policy Optimization (PPO) \cite{schulman2017proximal} algorithm to establish a learning baseline.

\textbf{Teleoperation and Behavior Cloning:} Expert recorded samples can help to investigate learning for challenging and long-horizon surgical tasks that are hard to solve with the current RL algorithms. We support the RoboMimic library \cite{mandlekar2021matters} which encompasses IL algorithms including behavior cloning (BC) and offline reinforcement learning (BCQ). \simName offers support for various I/O peripheral devices to teleoperate the dVRK gripper in simulation in real-time and record human expert data to train IL algorithms. Currently, our peripheral devices include keyboard, spacemouse, gamepad, VR controller, as well as the dVRK Master Tool Manipulator (MTM) to provide input commands and control the dVRK gripper in Cartesian space.

\section{Experimental Results}

In this section, we investigate task learning for the proposed surgical environments entirely in simulation and evaluate how well the trained policies transfer to a real world dVRK platform. To this end, we are answering the following questions:

\begin{enumerate}[noitemsep]
    \item How accurately do \simName's environments simulate physical interactions between the dVRK surgical robot and deformable objects?
    \item How well do RL algorithms perform for learning simple surgical subtasks?
    \item What is the benefit of the data collection from human-expert or code-generated trajectories for learning from demonstration for long-horizon tasks?
    \item How well do trained policies in \simName transfer from simulation environments to a physical dVRK surgical robot in the lab?
\end{enumerate}

\noindent \textit{Experimental Setup}:
The visualizations of experimentation environments are depicted in Fig. \ref{fig:fig1}. Each environment consists of either single or dual arms in a fixed location. The initial and desired states of tasks are randomized across the parallelized environments. The setup for the real world experiments consists of two arms mounted on a table. The surgical robot arms can be controlled either programmatically or through the MTM surgeon control platform.

\subsection{Reinforcement Learning}

\begin{figure}
    \centering
    \includegraphics[width=\linewidth]{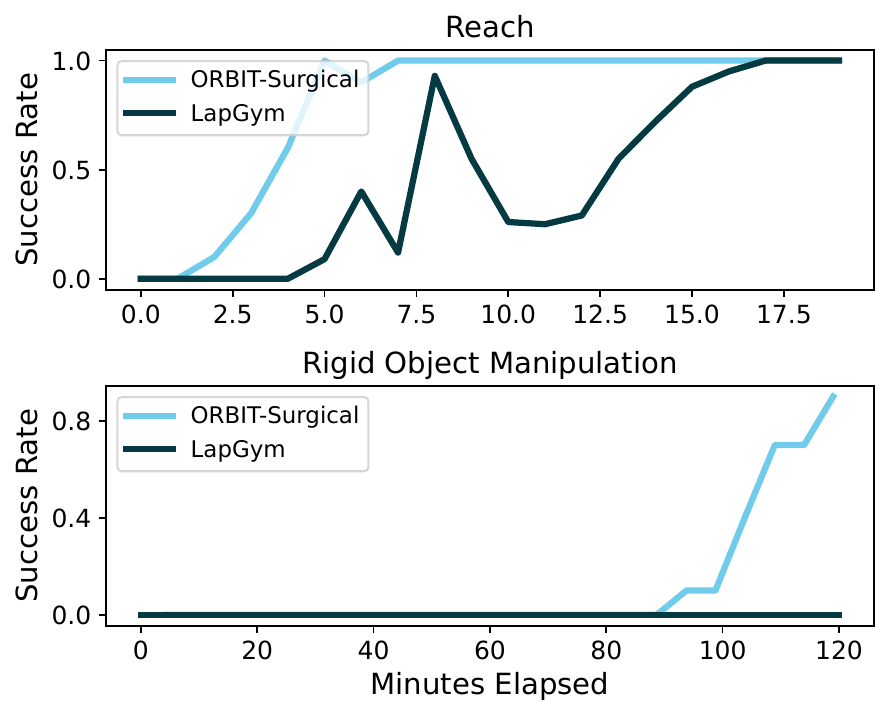}
    \caption{\textbf{Reinforcement Learning.} Success rate versus wall-clock time for RL policy training using the PPO algorithm on \texttt{Reach} task (above), and \texttt{Rigid Object Manipulation} task (below) evaluated for both \simName and LapGym across 10 trials for each saved checkpoint in simulation.
    }
    \label{fig:fig4}
\end{figure}

To demonstrate the high learning speed facilitated by \simName, RL policies were trained for surgical subtasks in \simName and were compared to those of LapGym \cite{scheikl2023lapgym}, a state-of-the-art surgical robotics simulator. Two tasks, i.e. \texttt{Reach} (Task1) and \texttt{Suture Needle Lift} (Task3) were selected for this comparison to study both simulators with and without object manipulation tasks. Both simulators have similar \texttt{Reach} tasks, however, the \texttt{Pick and Place} task from LapGym was selected to be of comparable complexity (i.e. requiring object interaction and data processing) with the \simName \texttt{Suture Needle Lift} task to study \texttt{Rigid Object Manipulation}. For the \texttt{Reach} task, both simulators learned the simple task quite fast since it does not involve interactions with objects. However, in \texttt{Rigid Object Manipulation} tasks, \simName started to learn to complete the task in around 100 minutes, whereas LapGym was not able to learn the task at all during the two-hour experiment time. Training rigid object manipulation tasks in \simName averaged 71700 FPS with RSL-RL while only getting 617 FPS throughput in LapGym with SB3 library on an NVIDIA 3090 RTX GPU. During training, \simName was able to parallelize 4096 environments. The success rate comparison is illustrated in Fig.~\ref{fig:fig4}.

\subsection{Behavior Cloning from Demonstration}

\begin{figure}
    \centering
    \includegraphics[width=0.95\linewidth]{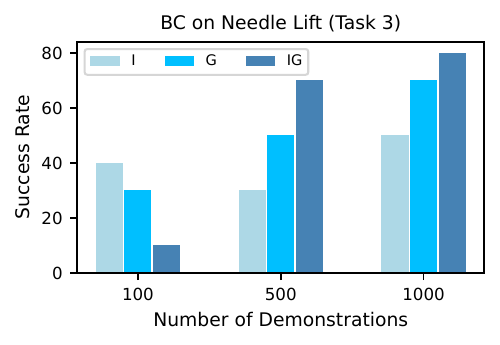}
    \caption{\textbf{Generalization of BC Policies in Simulation.} We collected demonstrations using hand-scripted state trajectories with variations in initial states and goal states for \texttt{Suture Needle Lift} (Fig.~\ref{fig:fig1}: Task3) in three different settings where only initial states are randomized (I), only goal states are randomized (G), and both initial and goal states are randomized (IG). Here, we compare the success rate of each trained policy across 10 trials in simulation. We note improved behavior for the policies trained on demonstrations with randomized initial and goal states.} 
    \label{fig:fig5}
\end{figure}

Surgical manipulation tasks involve a high level of precision. Current RL algorithms, at times, may not be capable of learning these tasks with optimal efficiency. In such instances, leveraging expert demonstrations for IL policies can be an effective strategy. \simName supports multiple teleoperation interfaces for data collection, including keyboard, VR controller, and MTM, where the collected data from an expert can be used to train BC models. Here, we evaluate the efficacy of BC policies for the single arm \texttt{Suture Needle Lift} task, where the initial position and goal states of the needle are either fixed or randomized. We collected 100, 500, and 1000 trajectories and subsequently trained BC policies. Due to the large number of demonstrations required for comparison, we collected demonstrations using hand-scripted state trajectories with variations in initial states and goal states. Fig.~\ref{fig:fig5} compares the success rate of policy execution for the \texttt{Suture Needle Lift} task in the simulation environment.

Long-horizon tasks are computationally expensive and can be hard to solve by plain BC algorithms. Instead, multi-stage BC policy can be learned efficiently, where long-horizon tasks are divided into subtasks, and sub-policies are trained to perform each subtask. All the subtasks are then chained for end-to-end policy execution. As an example, we show learning from demonstration for the \texttt{Needle Handover} task. We collected 1000 long-horizon trajectories of transferring a needle from one gripper to the other, and trained a single BC policy for the complete task which scored a 0 $\%$ success rate. We then divided the task into three stages, i.e. \texttt{Lift}, \texttt{Handover}, and \texttt{Reach} and trained BC policies to perform each subtask, respectively. All three subtasks were chained for end-to-end policy execution. For data collection, we recorded 100 trajectories for both \texttt{Lift} and \texttt{Handover}, and 300 trajectories for \texttt{Reach}. Policy execution for the \texttt{Needle Handover} task across 10 trials achieved a 40$\%$ success rate in simulation. Please visit the website for a successful needle handover policy execution in simulation.

\subsection{Synthetic Data Generation: Needle Segmentation}

\begin{figure}
    \centering
    \includegraphics[width=0.95\linewidth]{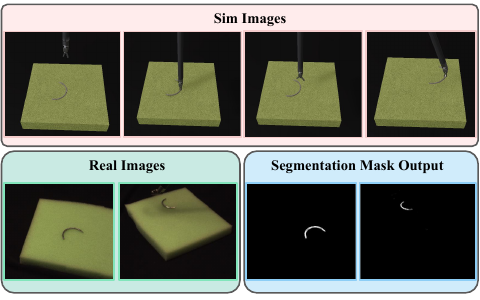}
    \caption{\textbf{Synthetic Data Generation.} Simulated images used as training data for needle segmentation (top). Real images of the workspace consisting of the needle are used for evaluation (bottom left). A model trained with both sim and real images is evaluated with a real dataset, resulting in masks shown on the bottom right.} 
    \label{fig:fig6}
\end{figure}

The fusion of photorealistic synthetic and real-world images enhances surgical tool segmentation models. 2900 $640 \times 480$ RGB images and corresponding segmentation masks rendered in the \simName simulator were paired with 100 RGB image-mask pairs of a needle in the real world, as shown in Fig.~\ref{fig:fig6}. Using this data in varying proportions, several models with the same hyperparameters were trained and compared in TABLE~\ref{table:table3}. Intersection over union (IoU) metric was considered to evaluate the performance of segmentation models on a separate evaluation set of 100 real-world RGB images by comparing the predicted segmentation to ground truth masks. The models trained exclusively on the simulation data generated segmentation masks that included other objects in the image such as the gripper, resulting in a low IoU. Combining simulation and real data resulted in predictions that do not select the gripper in the segmentation mask, while covering most of the needle area, leading to higher IoU scores than training exclusively on either simulation or real data.

\begin{table}
\vspace{-12pt}
\centering
\footnotesize
\caption{
\textbf{Real World Needle Segmentation Evaluation.} We note that dataset augmentation with simulation to a set of real images (50) improves segmentation IoU by over 2$\times$.
}
\begin{tabular}{ l |  l l c } 
\toprule
\rowcolor[HTML]{D7DAFF} 
 Training Regime & Sim Dataset Size & Real Dataset Size & IoU \\
\midrule
\midrule
 Sim Only & 2900 &  0 & 0.06\\
 \rowcolor[HTML]{EFEFEF} 
 Real Only & 0 &  100 & 0.15\\
 Sim+Real (20:1) & 1000 & 50 & \textbf{0.42}\\
 \rowcolor[HTML]{EFEFEF} 
 Sim+Real (29:1) & 2900 &  100 & \textbf{0.40}\\ 
\bottomrule
\end{tabular}
\label{table:table3}
\end{table}

\subsection{Sim-to-Real: Deployment on the dVRK}

We developed state machine robot controllers for two tasks involving deformable objects, \texttt{Threaded Needle Pass Ring} and \texttt{Shunt Insertion}, and deployed the trajectories recorded in the simulation environment on a real-world dVRK platform.

\begin{figure}[!t]
    \centering
    \includegraphics[width=0.9\linewidth]{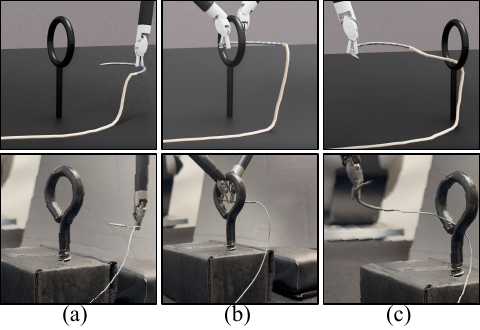}
     \caption{\textbf{Sim-to-Real.} \texttt{Threaded Needle Pass Ring} task in \simName (top) and on a real physical dVRK robot (bottom). (a) Right gripper grasps and lifts a threaded needle, (b) right gripper hands over the needle to the left gripper through the ring, (c) handover is complete and the left gripper moves away to a desired final location. Note how suture thread deformation in sim closely matches that in real.}
    \label{fig:fig7}
\end{figure}

For the \texttt{Threaded Needle Pass Ring} task, we designed an environment containing two dVRK arms, a needle attached to a thread as well as a loop on a pole. We programmed the state machine to grasp the needle using the right dVRK arm, rotate it through the ring and regrasp the needle with the left dVRK arm as seen in Fig. \ref{fig:fig7}. The history of joints and gripper jaw states for both arms were recorded in \simName and transferred to the real robot. The joints and gripper commands were adapted to match the real robot position offset and jaw command range. We used a surgical suture with a GS-21 needle for the real setup and replayed the trajectory on both arms. A qualitative comparison is shown in Fig.~\ref{fig:fig7}.

The \texttt{Shunt Insertion} task consists of two horizontal flexible tubes of equal length but different diameters and a single dVRK arm. This experiment shows the capacity to simulate shunt insertion maneuvers. The smaller diameter, more flexible tube was grasped by the surgical robot arm and inserted into the stiffer, larger diameter tube as shown in Fig.~\ref{fig:fig1}: Task 12. Similar to the \texttt{Threaded Needle Pass Ring} experiment, the joints trajectory and gripper states were recorded in \simName and replayed on the real robot. For the real setup, we used latex surgical tubing for the shunt and a clear vinyl tube for the artery. The artery was held in place using a fixed clamp. Please refer to the video for real and simulated robot experiments.

\begin{figure}
    \centering
    \includegraphics{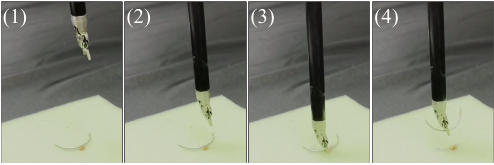}
     \caption{\textbf{Zero-shot \texttt{Suture Needle Lift} Policy Transfer.} (1) Starting position of the dVRK arm and suture needle, (2) arm moving towards the needle, (3) grasping the needle, (4) successfully lifting the needle. The policy was fully trained in \simName and deployed on the physical robot.}
    \label{fig:fig8}
\end{figure}

We deployed a PPO policy trained with RSL-rl on the \texttt{Suture Needle Lift} task in Cartesian space on a real dVRK setup. The needle was placed below one of the grippers and on top of a green piece of foam for safety reasons shown in Fig.~\ref{fig:fig8}. All of the real-world observations for this task were derived from the real dVRK's joint encoders and other known values except for the needle position which was obtained manually. The initial needle position was measured and hard coded in the observation space. When the gripper grasped the needle, we used the last known relative pose between the gripper and the needle along with the end-effector position to derive the current needle position. We performed 20 trials, and observed that the transferred policy achieved a 50\% success rate. One execution run of the policy is shown in Fig.~\ref{fig:fig8}.

\section{Conclusions and Future Work}

This paper introduces \simName, an open physics-based simulator which provides contact-rich surgical simulation environments for robot learning. GPU parallelization enables \simName to train reinforcement learning (RL) policies for surgical robotics tasks on the order of hours. Through multiple teleoperation interfaces, \simName leverages real-world demonstrations for policy learning in simulation. \simName enables high-fidelity synthetic data generation by harnessing GPU-accelerated physics with ray-traced rendering. Together with high-fidelity rendering and high-speed simulation, we hope to accelerate progress in learning-based assistive autonomy in robotic surgery.

While \simName provides high throughput in state-based environments, vectorized image rendering is still under development. This will enable us to incorporate visual inputs into training RL policies. \simName also does not model the cable stretching effects found on real dVRK surgical robots, as demonstrated by the success rate of 50\% when deploying the RL policy for the \texttt{Suture Needle Lift} task. In future work, we will incorporate cutting of soft material~\cite{heiden2023disect} and provide algorithms to solve longer-horizon end-to-end tasks such as suturing and knot tying.

\clearpage

\bibliographystyle{IEEEtran}
\bibliography{orbit-surgical}

\end{document}